\documentclass[journal]{IEEEtran}

\usepackage{cite}

\ifCLASSINFOpdf
   \usepackage[pdftex]{graphicx}
   \graphicspath{{../pdf/}{../jpeg/}}
   \DeclareGraphicsExtensions{.pdf,.jpeg,.png}
\else
   \usepackage[dvips]{graphicx}
   \graphicspath{{../eps/}}
   \DeclareGraphicsExtensions{.eps}
\fi

\usepackage{amsmath}

\usepackage{algorithmic}

\usepackage{array}

\ifCLASSOPTIONcompsoc
  \usepackage[caption=false,font=normalsize,labelfont=sf,textfont=sf]{subfig}
\else
  \usepackage[caption=false,font=footnotesize]{subfig}
\fi
\usepackage{url}

\usepackage{makecell}

\begin{document}
\title{CAD-Net: A Context-Aware Detection Network for Objects in Remote Sensing Imagery}

\author{Gongjie~Zhang, Shijian~Lu and Wei~Zhang
}


\maketitle

\begin{abstract}
Accurate and robust detection of multi-class objects in optical remote sensing images is essential to many real-world applications such as urban planning, traffic control, searching and rescuing, etc. However, state-of-the-art object detection techniques designed for images captured using ground-level sensors usually experience a sharp performance drop when directly applied to remote sensing images, largely due to the object appearance differences in remote sensing images in term of sparse texture, low contrast, arbitrary orientations, large scale variations, etc. This paper presents a novel object detection network (CAD-Net) that exploits attention-modulated features as well as global and local contexts to address the new challenges in detecting objects from remote sensing images. The proposed CAD-Net learns global and local contexts of objects by capturing their correlations with the global scene (at scene-level) and the local neighboring objects or features (at object-level), respectively. In addition, it designs a spatial-and-scale-aware attention module that guides the network to focus on more informative regions and features as well as more appropriate feature scales. Experiments over two publicly available object detection datasets for remote sensing images demonstrate that the proposed CAD-Net achieves superior detection performance. The implementation codes will be made publicly available for facilitating future researches. 
\end{abstract}


\IEEEpeerreviewmaketitle

\section{Introduction} \label{section:1}
\IEEEPARstart{T}{he} recent advances in satellites and remote sensing technologies have been leading to a huge amount of high-definition remote sensing images every day that simply goes beyond any manual manipulation and processing. Automated analysis and understanding for remote sensing images have therefore become critically important to make these images useful in many real-world applications such as urban planning, searching, rescuing, environmental monitoring, etc. In particular, multi-class object detection, which simultaneously localizes and categories various objects (e.g., planes, vehicles, bridges, roundabouts, etc.) within remote sensing images, has become possible due to the increase of sensor resolutions. This challenge goes beyond the traditional scene-level analytics \cite{Jiayuan,Tao,TGRS-Scene} that aim to identify the scene semantics of remote sensing images such as building, grassland, sea, etc., and it has attracted increasing research interest in recent years.

\begin{figure}[!t]
\begin{center}
   \includegraphics[height=121mm,width=1\linewidth]{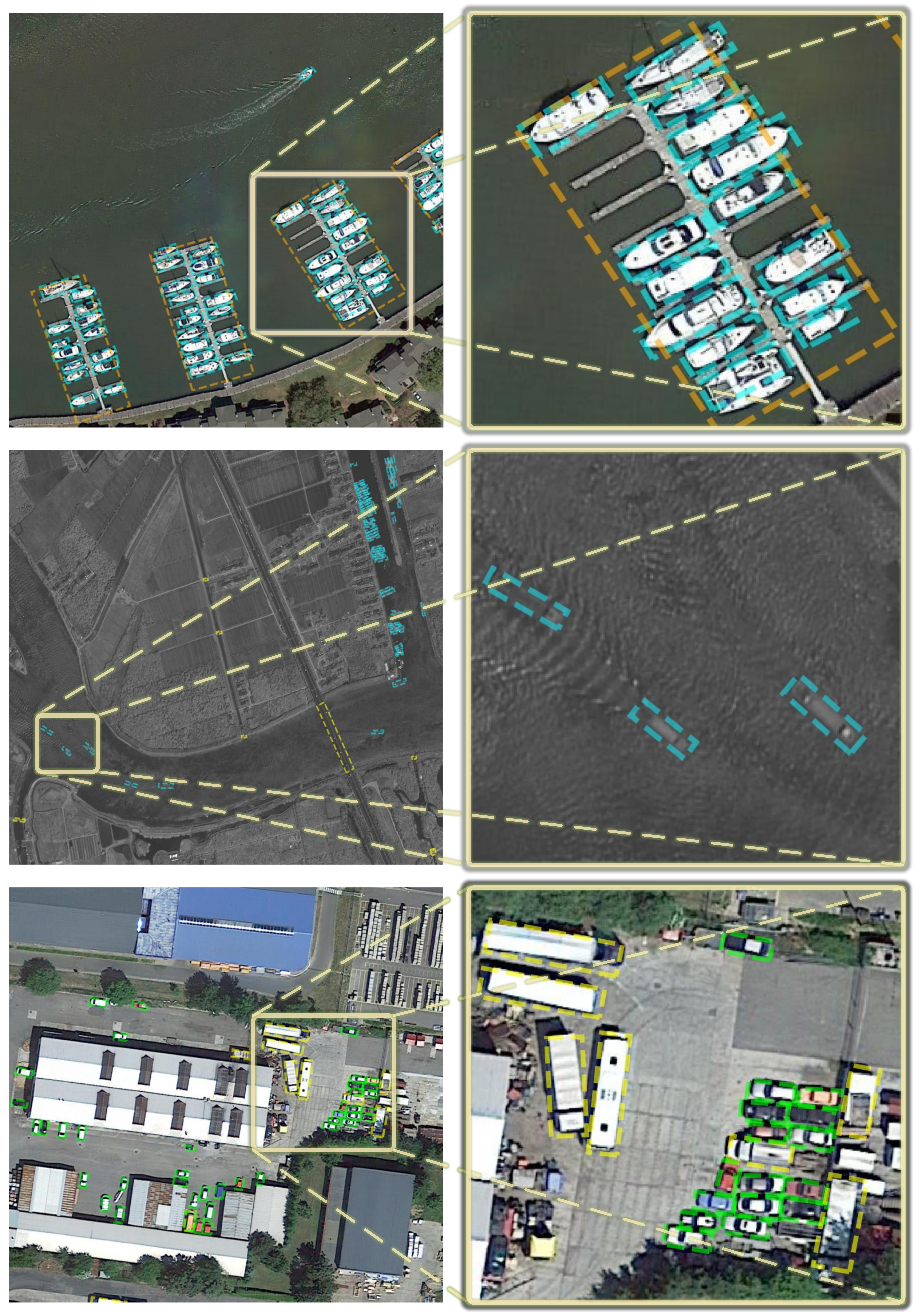}
\end{center}
   \caption{Illustration of results produced by the proposed CAD-Net for object detection in optical remote sensing imagery: Multi-class objects with different types of image degradation and information loss in colors, contrast and texture are detected and recognized correctly.}
\label{fig:DetectionExample}
\end{figure}

The fast development of deep neural networks especially convolution neural networks (CNNs) has raised the bar of object detection greatly in recent years. A number of CNN-based object detectors \cite{RCNN, FastRCNN, FasterRCNN, SSD, YOLO, YOLO9000, RFCN} have been proposed and very promising results have been achieved over several large-scale object detection datasets such as PASCAL VOC \cite{POSCAL_VOC_CHALLENGE} and MS COCO \cite{MSCOCO_CHALLENGE}. On the other hand, most existing techniques often experience a sharp performance drop while applied to remote sensing images~\cite{DOTA}, largely due to three factors as illustrated in Fig.~\ref{fig:DetectionExample}. First, objects in optical remote sensing images usually lack visual clues such as image contrast and texture details that are critically important to the performance of state-of-the-art detection techniques. Second, objects in remote sensing images are usually densely distributed, appear in arbitrary orientations and have large scale variations, which make object detection an even more challenging task. Third, objects captured in optical remote sensing images usually suffer from a large amount of noises due to various interference while light gets reflected and travels a long way back to satellite sensors.

\begin{figure*}[t]
\begin{center}
   \includegraphics[height=75mm,width=1\linewidth]{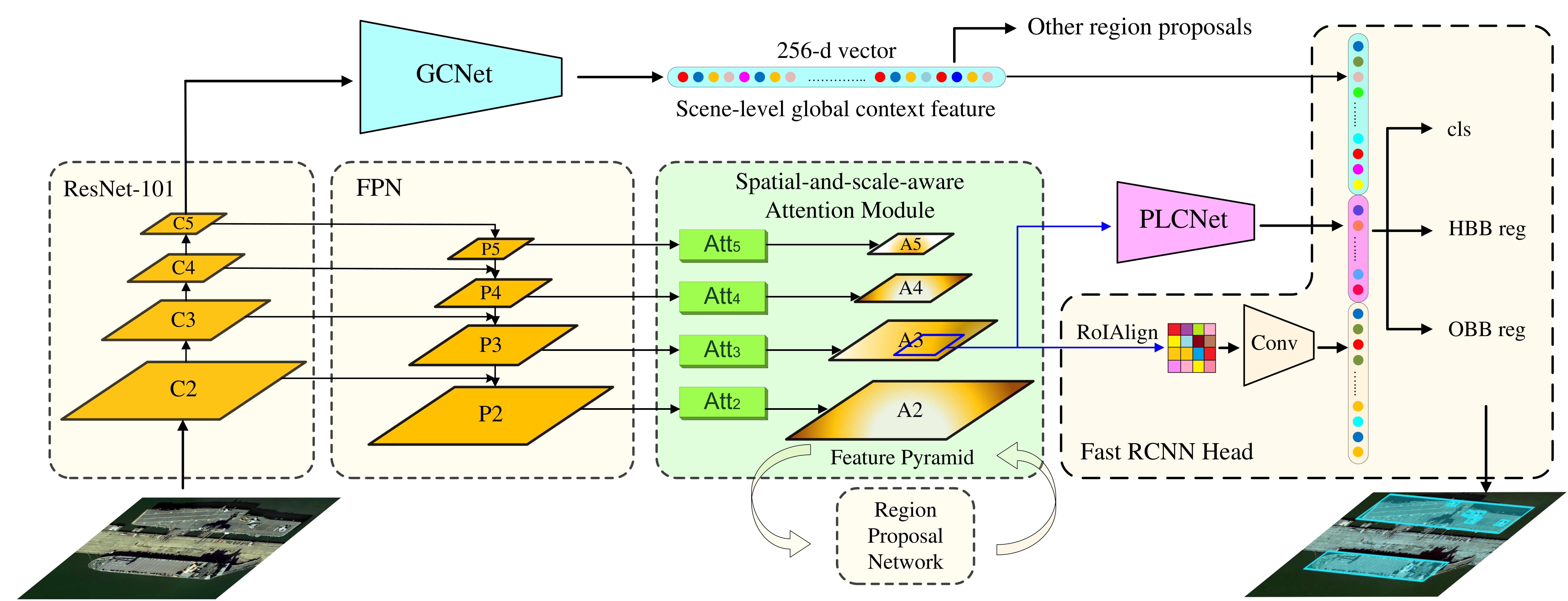}
\end{center}
   \caption{The framework of the proposed CAD-Net: Leveraging the Faster RCNN with feature pyramid networks (in beige), a global context network (GCNet highlighted in cyan) and a pyramid local context network (PLCNet highlighted in purple) are designed to learn global contexts at scene level and local contexts at object level, respectively. An spatial-and-scale-aware attention module (in light green) is designed to guides the network to focus on more informative regions at the appropriate feature scales while suppressing irrelevant information. In addition to standard horizontal bounding box (HBB) regression, an oriented bounding box (OBB) regression branch is added to produce OBB results which better align with arbitrary oriented attributes of objects in remote sensing images.}
\label{fig:NetworkOverview}
\end{figure*}

In this work, we design a Context-Aware Detection Network (CAD-Net) for object detection in optical remote sensing images. Fig.~\ref{fig:NetworkOverview} shows the overview of the proposed CAD-Net. As Fig.~\ref{fig:NetworkOverview} shows, the CAD-Net consists of a Global Context Network (GCNet) that learns the correlation between interested objects and their corresponding global scenes, i.e., the correlation between features of objects and features of the whole image. The GCNet is inspired by the observations that optical remote sensing images usually cover large areas where the scene-level semantics often provides important clues on both object locations and object categories, e.g. ships often appear in seas/rivers, helicopters hardly appear around residence areas, etc. In addition, the CAD-Net consists of a Pyramid Local Context Network (PLCNet) that learns multi-scale co-occurrence features and/or co-occurrence objects surrounding the objects of interest. Compared with images captured by ground-level sensors, remote sensing images from the top view often contain richer and more distinguishable co-occurrence features and/or objects that are very useful for object category and object position reasoning, e.g., vehicles appearing around each other, ships in harbors, bridges above rivers, etc. Further, a spatial-and-scale-aware attention module is designed to guide the network focus on more informative contextual regions at the right image scales.

The contributions of this work are fourfold. First, it designs an innovative context-aware network to learn global and local contexts for optimal object detection in optical remote sensing images. To the best of our knowledge, this is the first work to incorporate global and local contextual information for object detection in remote sensing images. Second, it designs a spatial-and-scale-aware attention module that guides the network to focus on more informative regions at the appropriate image feature scales. Third, it verifies the uniqueness of remote sensing object detection, and also provides an insightful and novel solution to bridge the gap with respect to object detection from images captured by ground-level sensors. Fourth, without bells and whistles, it develops an end-to-end trainable detection network that obtains state-of-the-art performance over two challenging object detection datasets for remote sensing images.

The rest of this paper is organized as follows. Related works are first described in the following Section II. The proposed method including the contextual networks and the spatial-and-scale-aware attention is then presented in Section III in details. Section IV further presents implementation details and experimental results. Some concluding remarks are finally summarized in Section V.

\section{Related Work} \label{section:2}

\subsection{Generic Object Detection}

The performance of object detection has been improved greatly with the recent advances in deep neural networks. State-of-the-art object detection techniques can be broadly classified into two categories, namely, two-stage detectors and single-stage detectors. Two-stage detectors mainly refer to Region-based Convolutional Neural Networks (R-CNN), including the original R-CNN~\cite{RCNN} that leverages Selective Search~\cite{SelectiveSearch} to generate region proposals and further CNNs for feature extraction from each proposal, Fast RCNN~\cite{FastRCNN} that improves R-CNN by sharing computation of feature extraction for all region proposals, Faster RCNN~\cite{FasterRCNN} that proposes a novel Region Proposal Network (RPN) for proposal generation and the very recent Feature Pyramid Network (FPN)~\cite{FPN} that builds a feature pyramid to learn features of different scales to deal with large object scale variation. Single-stage detectors such as YOLO~\cite{YOLO, YOLO9000} and SSD~\cite{SSD}, which can perform nearly real-time object detection, do not require proposal generation procedure. On the other hand, they usually have lower detection accuracy than two-stage detectors, especially in the presence of a large amount of small object instances. These generic objects detection methods have been modified for many specific tasks~\cite{MaskRCNN, R2CNN, MaskTextSpotter, Azimi_DOTA}, and have achieved very promising results.

Our proposed CAD-Net is based on the widely used two-stage detector--Faster RCNN~\cite{FasterRCNN}. It instead designs some innovative components to address the specific characteristics of objects within remote sensing images, such as lack of image contrast and texture details, arbitrary object orientations, etc, more details to be shared in the ensuing Section III.

\subsection{Object Detection in Optical Remote Sensing Images}
Detecting objects in remote sensing images has been studied for years. Most works \cite{cheng2014multi, RSD1, RSD4, RSD3, RSD5, RSD6, RSD2, Huili} in earlier days use hand-crafted features, producing very limited detection performance. In addition, most of them were specifically developed for a single type of objects, and often fail while dealing with objects with a cluttered background. In recent years, Convolutional Neural Networks (CNNs) have been exploited for object detection in remote sensing images and promising results have been obtained~\cite{RSDD3,RSDD1,RSDD4,RSDD2,YangXue_ShipDetection,Azimi_DOTA}. For example, \cite{RSDD1} uses multi-scale CNN features for airport detection, \cite{RSDD2} takes a weakly supervised learning framework for aircraft detection using CNNs, \cite{RSDD4} and \cite{YangXue_ShipDetection} integrate Dense Feature Pyramid Network (DFPN) and rotation regression for ship detection in remote sensing images, and \cite{RSDD3} uses CNN features for multi-class object detection in remote sensing images. 

Our proposed CAD-Net is a multi-class detector that can detect many different types of objects within the same image in one go. It exploits deep neural networks for optimal object detection performance in optical remote sensing images. In addition, it is evaluated over large-scale datasets that can be easily accessed through the Internet.

\subsection{Contexts and Attention Mechanism}

The usefulness of contexts in image understanding has been verified in both psychological and empirical studies \cite{VisualContext_Nature, EmpiricalStudyContext}. For object detection, several studies \cite{InsideOutsideNet, FindingTinyFaces, ContextLocNet} have shown that contextual information can help improve the object detection performance, especially in the presence of small objects. In addition, the mechanism of visual attention~\cite{MachineTranslationAttention} has shown its usefulness in many computer vision tasks by guiding the processing to more informative and relevant regions. For example, \cite{Xu_ImageCaption_Attention} uses visual attention to determine the most relevant regions in image captioning, \cite{Zhu_VQA_Attention} uses visual attention for feature fusion for Visual Question Answering, \cite{SCACNN} combines spatial attention and channel-wise attention to determine selective regions of selective categories, and \cite{YK_attention_PersonReID} uses global and local attention for person re-identification.

Our proposed CAD-Net fuses contextual information and attention mechanism for optimal object detection performance in remote sensing images. Contextual information is adopted to provide extra guidance for objects with few visual clue and low contrast; while spatial-and-scale-aware attention is designed for better robustness to scale variance and noises. Extensive experiments in Section \ref{section:4} demonstrate their effectiveness.

\section{Proposed Method} \label{section:3}

The framework of our proposed Context-Aware Detection Network (CAD-Net) is illustrated in Fig.~\ref{fig:NetworkOverview}. As the figure shows, our CAD-Net is built on the structure of classical two-stage detection network--Faster RCNN \cite{FasterRCNN} with FPN~\cite{FPN}. A Global Context Network (GCNet) and a Pyramid Local Context Network (PLCNet) are designed and fused to extract contextual information at global scene level and local object level, respectively. A spatial-and-scale-aware attention module is designed which guides the network to focus on more informative regions as well as more appropriate image feature scales. All designed components are off-the-shelf and can be incorporated into existing detection networks without any adaptation and extra supervision information. More details are to be discussed in the ensuing subsections.

\subsection{Leveraging Contextual Information}

Given an image $\mathcal{I}$ and a region proposal $\mathcal{P}$, the detection of the objects $\mathcal{O_P}$ with respect to $\mathcal{P}$ can be formulated by:
\begin{equation}\label{equ:det1}　
\mathcal{O_P} = Det(\mathcal{I}, \mathcal{P})
\end{equation}
\noindent where $Det(\cdot)$ denotes joint object classification and bounding box regression. In the widely adopted region-based detection approaches, Eq.~\ref{equ:det1} is often approximated by RoIPooling~\cite{FastRCNN} that guides the network to focus on the proposal region and ignore the rest parts of the image. The new formulation can thus be presented by:
\begin{equation}\label{equ:det2}　
\mathcal{O_P} = Det[\Psi(\mathcal{P}, \mathcal{I}), \mathcal{P}]
\end{equation}
\noindent where $\Psi(\cdot)$ denotes the RoIPooling operation.

The approximation in Eq.~\ref{equ:det2} is based on the assumption that all useful information for a specific region $\mathcal{P}$ lies within the region itself. This assumption works for most images from ground-level sensors where discriminative object features are usually captured and kept well. But for optical remote sensing images, discriminative object features such as edges and texture details are often severely degraded due to various noises and information loss. Under such circumstance, global and local contexts that are often correlated with objects of interest closely become important and should be incorporated to compensate the feature degradation and information loss. The incorporation of the global and local contexts can thus be formulated as follows:
\begin{equation}\label{equ:det3}　
\mathcal{O_P} = Det\left\{[\Psi(\mathcal{P}, \mathcal{I}) ; G(\mathcal{I}) ; L(\mathcal{P}, \mathcal{I}) ], \mathcal{P} \right\}
\end{equation}
\noindent where $G(\cdot)$ denotes GCNet for obtaining global contextual features, $L(\cdot)$ denotes PLCNet for obtaining local contextual features and $(\cdot ; \cdot)$ denotes concatenation.

\medskip
\subsubsection{Global Context Network}
Remote sensing images usually capture a large area of land that carries strong semantic information that characterizes the captured scene. In addition, the semantics of the captured scenes are often closely correlated with objects within the scenes, e.g. sea versus ship, airport versus airplane, etc. With these observations, we design a Global Context Network (GCNet) that learns the global scene semantics and uses them as certain priors for better detection of objects in remote sensing images. More specifically, the GCNet learns the correlation between a scene and objects within the scene and use the learned correlation as certain global contexts to compensate the loss of discriminative object features. The GCNet can be formulated as follows:
\begin{equation}\label{equ:gcnet}　
G(\mathcal{I}) = \psi\left\{\Phi_G\left[\Lambda( \mathcal{I} )\right]\right\}
\end{equation}
\noindent where $\Lambda( \mathcal{I} )$ denotes the final feature maps of the feature extraction network, i.e. the C5 level of the ResNet-101 backbone as illustrated in Fig.~\ref{fig:NetworkOverview}, $\Phi_G(\cdot)$ is implemented by a stack of convolutional layers that extracts global features, and $\psi(\cdot)$ denotes a pooling operation that squeezes the spatial channels of the feature maps into a vector which helps to suppress the sensitivity to scale variations. We empirically adopt global average pooling for $\psi(\cdot)$ in our implemented system.

\begin{figure}[t] 
\begin{center}
   \includegraphics[height=75mm, width=1.0\linewidth]{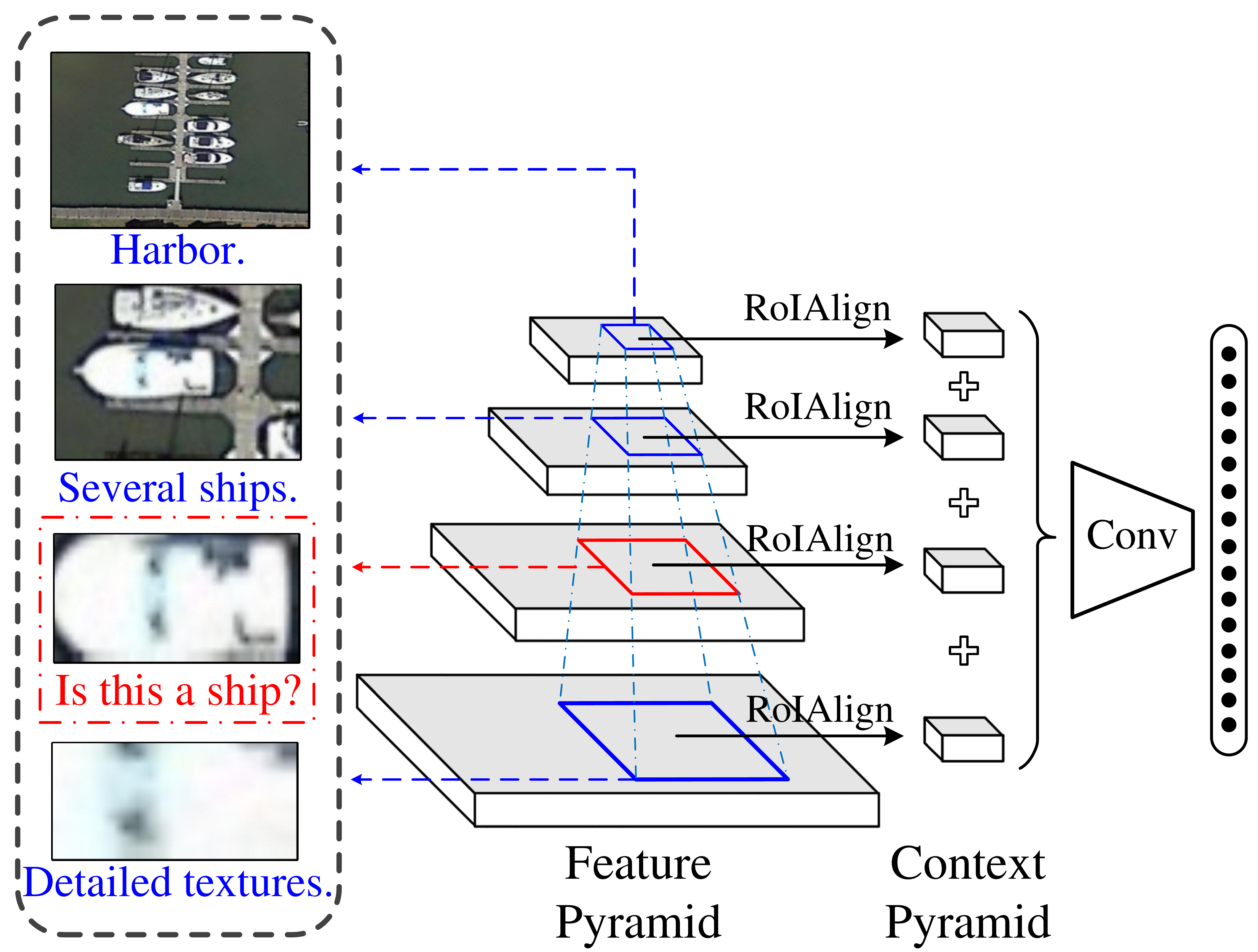}
\end{center}
   \caption{Illustration of the framework of our proposed  Pyramid  Local  Context  Network (PLCNet): The PLCNet is able to extract features of the proposal region from each feature scale, and learn their correlations to serve as supplementary information for detection.}
\label{fig:PLCNet}
\end{figure}

\medskip
\subsubsection{Pyramid Local Context Network} 
Besides global contexts, local contexts which characterize the correlation between an object and its neighboring objects and/or features also capture useful information and can be exploited to compensate the information loss. Based on the observation that both objects and their local contexts are scale sensitive, we design a Pyramid Local Context Network (PLCNet) to learn the object/feature correlation between objects and their local contexts as illustrated in Fig.~\ref{fig:PLCNet}. 

\begin{figure}[t]
\begin{center}
   \includegraphics[height=75mm, width=1\linewidth]{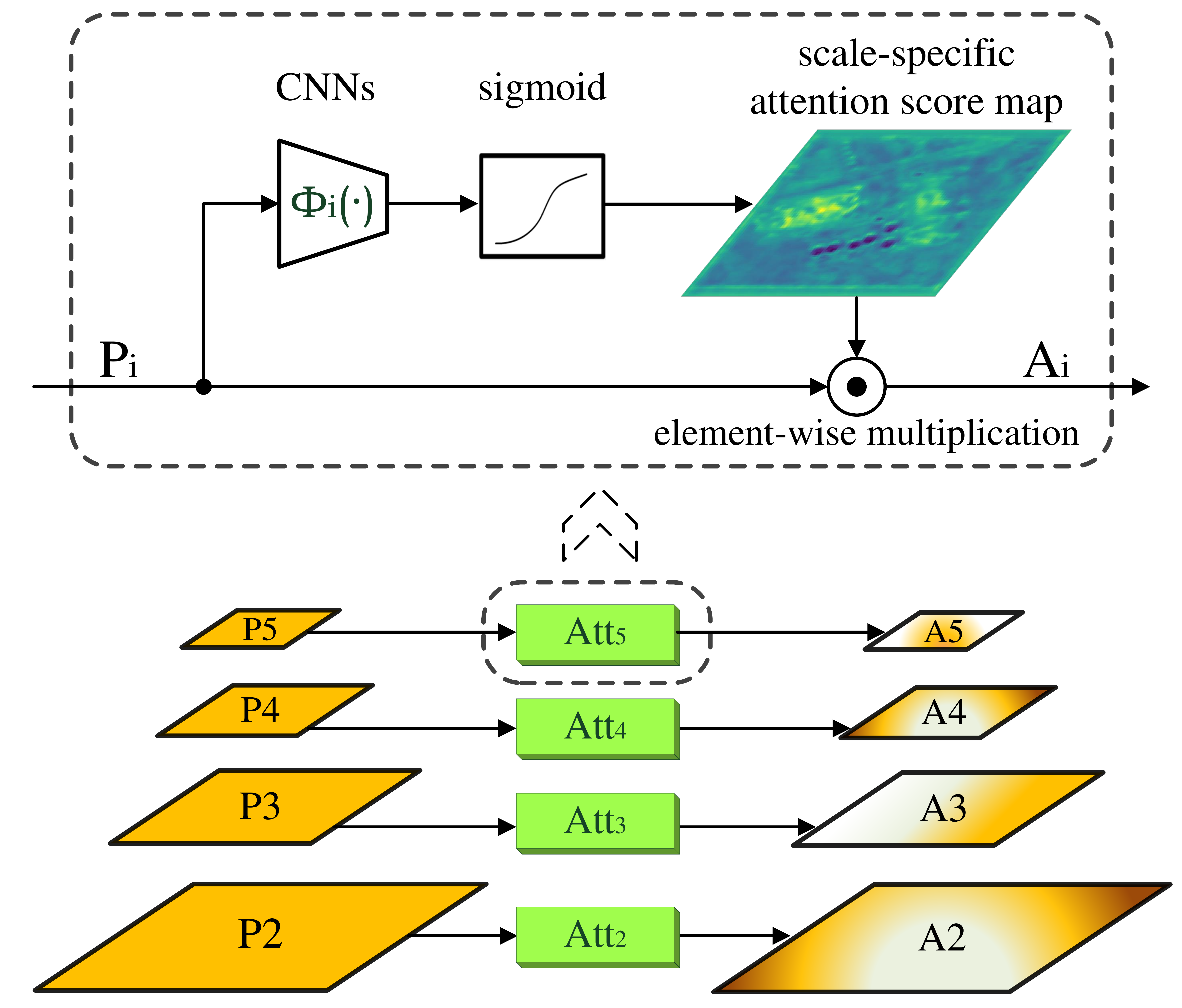}
\end{center}
   \caption{Illustration of our proposed spatial-and-scale-aware attention module: scale-aware attention is applied at different network layers to focus on more informative regions at different image scales.}
\label{fig:AttenMechanism}
\end{figure}

Given a region proposal $\mathcal{P}$ (e.g. the ship proposal in red-color box in Fig.~\ref{fig:PLCNet}), a set of local contexts of the corresponding region of different scales are employed to learn the cross-scale local contexts around $\mathcal{P}$ as illustrated in Fig.~\ref{fig:PLCNet}. A \textbf{Context Pyramid} is designed, which first extracts and concatenates pooled features at different scales and then fuses the concatenated features through convolution (i.e. \textit{Conv} in Fig.~\ref{fig:PLCNet}). The fused features are finally concatenated with the region features as well as the aforementioned global context features for proposal classification and bounding box regression.

As illustrated in Fig.~\ref{fig:PLCNet}, even humans will find it challenging to tell whether the proposed region (highlighted in red-color box) is a ship by just focusing on the proposed region itself. Under such circumstance, the local contexts from different scales (e.g. cluster of ships and harbors shown in Fig.~\ref{fig:PLCNet}) will provide strong clues that the region proposal is very likely to be a ship. The PLCNet is trained to learn such correlated features and/or objects, which often help a lot in the presence of sparse texture, low contrast as well as severe information loss in optical remote sensing images.

\begin{figure*}[t!] 
\begin{center}
   \includegraphics[height=70mm, width=0.9\linewidth]{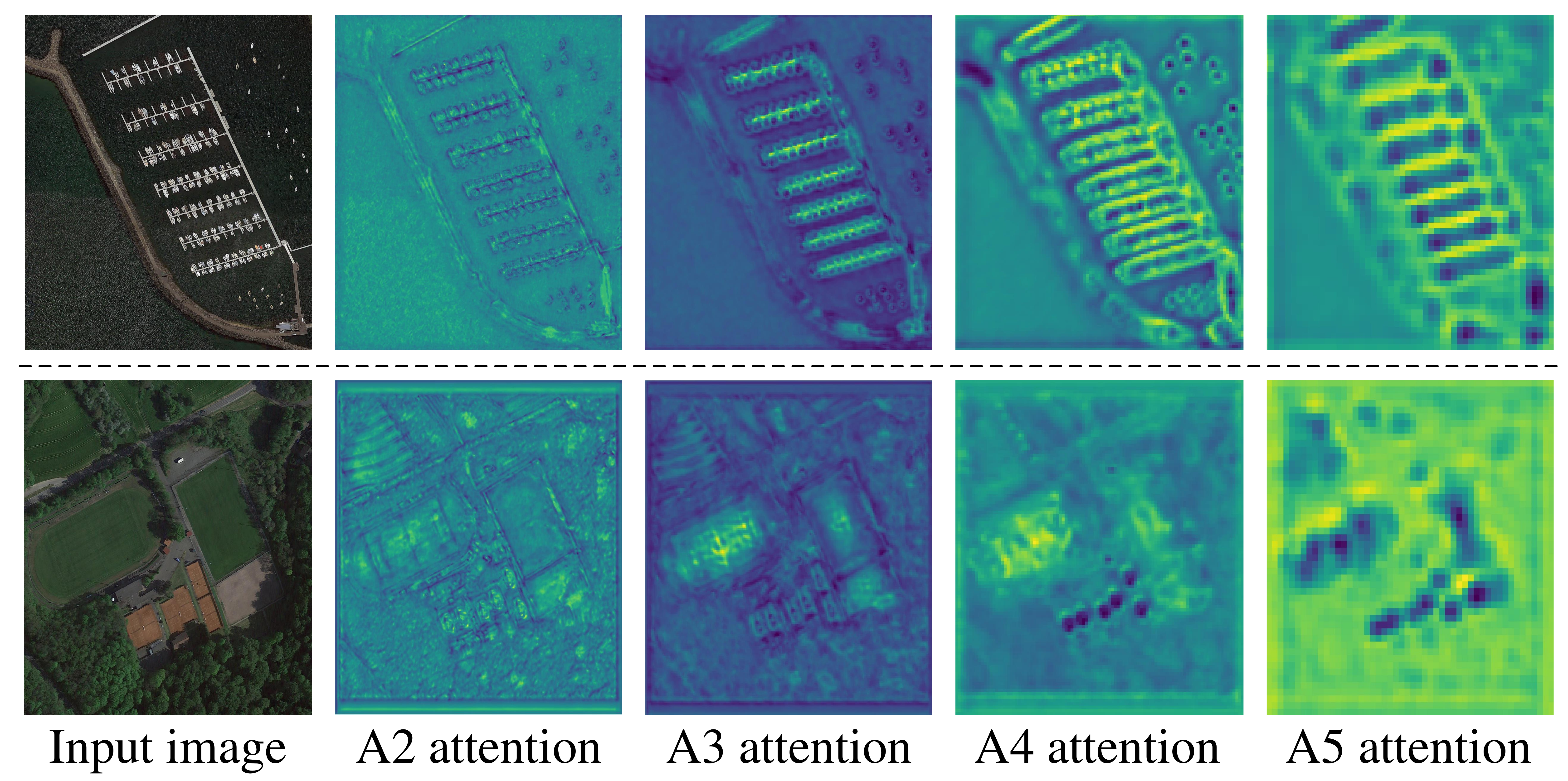}
\end{center}
   \caption{Illustration of our proposed spatial-and-scale-aware attention responses at different feature scales. Brighter regions indicate higher attention responses. The proposed spatial-and-scale-aware attention module is capable of focusing on informative regions at the appropriate feature scales, while suppressing irrelevant and noisy areas.}
\label{fig:AttenMap}
\end{figure*}

\subsection{Spatial-and-Scale-Aware Attention Module}

Visual attention has been proven very useful in different computer vision tasks such as image captioning, scene text recognition, etc. The idea is inspired by the human vision system that does not process an entire image at one go but tends to focus on more informative regions sequentially. In this work, we design a spatial-and-scale-aware attention module that learns to adaptively focus on more prominent regions (spatial-aware) at relevant scales of feature maps (scale-aware). The spatial-aware feature helps the network to handle objects with sparse texture and low contrast with background, and the scale-aware feature helps to handle objects of very different scales. The combination of both facilitate the learning of object detection models for remote sensing images.

The proposed spatial-and-scale-aware attention module is built upon the FPN-generated feature pyramid that extracts feature maps $P_2-P_5$ as illustrated in Fig.~\ref{fig:AttenMechanism}. For feature of one specific scale $P_i (i \in [2,5])$, an attention-modulated feature map is determined as follows:

\begin{equation}\label{equ:atten1}　
S_i = \sigma[ \Phi_i(P_i) ]
\end{equation}
\begin{equation}\label{equ:atten2}　
A_i = S_i \odot P_i
\end{equation}

\noindent where $\sigma(\cdot)$ is a sigmoid function, $S_i$ is the attention map of the i-th feature map, $A_i$ is the i-th attention-modulated feature map, and $\odot$ denotes element-wise multiplication. The attention map computation $\Phi_i(\cdot)$ is achieved by a stack of convolutional layers. Note that a separate $\Phi_i(\cdot)$ is implemented to compute each scale-specific attention map. This design ensures that our proposed attention module is both spatial-aware and scale-aware, enabling it to focus on more informative regions at appropriate scales with irrelevant information suppressed.

Fig.~\ref{fig:AttenMap} shows the attention response maps that are generated by the proposed spatial-and-scale-aware attention module. As Fig.~\ref{fig:AttenMap} shows, our proposed attention module is not only spatial aware, but also scale-aware, which can selectively focus on more informative regions for features of different scales. For example, small-scale ships get stronger responses at lower network layers A2 and A3 (as shown in Fig.~\ref{fig:AttenMechanism}) that capture more detailed information, while large-scale harbors get stronger responses at deeper network layers A4 and A5 that capture more high-level information as illustrated in the first sample image. In addition, our attention module is able to guide the network to focus on useful texture details that are degraded by noises, e.g. the skeleton of the harbors in the first sample image and the middle line of the court in the second sample image.

\begin{table*}[t!]
\begin{center}
\caption{Experiment Results of CAD-Net as well as Comparisons with State-of-the-art Techniques over the DOTA dataset test set. Methods specifically designed for remote sensing images are marked with $\triangle$ for fair comparison.}
\label{tab:dota_results}
\resizebox{\textwidth}{0.15mm}{
\begin{tabular}[t]{ccccccccccccccccccc}
\hline
{\bf method }&  {\bf data} & {\bf mAP} & {\bf PL} & {\bf BD} & {\bf bridge} & {\bf GTF} & {\bf SV} & {\bf LV} & {\bf ship} & {\bf TC} & {\bf BC} & {\bf ST} & {\bf SBF} & {\bf RA} & {\bf harbor} & {\bf SP} & {\bf HC} \\
\hline
\hline
YOLO\_v2~\cite{YOLO9000,DOTA} & T+V & 25.5 & 52.8 & 24.2 & 10.6 & 35.5 & 14.4 & 2.4 & 7.4 & 51.8 & 44.0 & 31.4 & 22.3 & 36.7 & 14.6 & 22.6 & 11.9 \\
\hline
SSD~\cite{SSD,DOTA} & T+V & 17.8 & 41.1 & 24.3 & 4.6 & 17.1 & 15.9 & 7.7 & 13.2 & 40.0 & 12.1 & 46.9 & 9.1 & 30.8 & 1.4 & 3.5 & 0.0 \\
\hline
R-FCN~\cite{RFCN,DOTA} & T+V & 30.8 & 39.6 & 46.1 & 3.0 & 38.5 & 9.1 & 3.7 & 7.5 & 42.0 & 50.4 & 67.0 & 40.3 & 51.3 & 11.1 & 35.6 & 17.5 \\
\hline
FR-H~\cite{FasterRCNN,DOTA} & T+V & 40.0 & 49.7 & 64.2 & 9.4 & 56.7 & 19.2 & 14.2 & 9.5 & 61.6 & 65.5 & 57.5 & 51.4 & 49.4 & 20.8 & 45.8 & 24.4 \\
\hline
FR-O~\cite{FasterRCNN,DOTA} & T+V & 54.1 & 79.4 & 77.1 & 17.7 & 64.1 & 35.3 & 38.0 & 37.2 & 89.4 & 69.6 & 59.3 & 50.3 & 52.9 & 47.9 & 47.4 & 46.3 \\
\hline
$\triangle$ R-DFPN~\cite{RSDD4,YangXue_ShipDetection} & T+V & 57.9 & 80.9 & 65.8 & 33.8 & 58.9 & 55.8 & 50.9 & 54.8 & 90.3 & 66.3 & 68.7 & 48.7 & 51.8 & 55.1 & 51.3 & 35.9  \\
\hline
$\triangle$ Yang et al.~\cite{YangXue_ShipDetection} & T+V & 62.3 & 81.3 & 71.4 & 36.5 & 67.4 & 61.2 & 50.9 & 56.6 & 90.7 & 68.1 & 72.4 & 55.1 & 55.6 & 62.4 & 53.4 & 51.5  \\
\hline
$\triangle$ Azimi et al.~\cite{Azimi_DOTA} & T & 65.0 & 81.2 & 68.7 & 43.4 & 61.1 & 65.3 & 67.7 & 69.2 & 90.7 & 71.5 & 70.2 & {\bf 55.4} & 57.3 & 66.5 & 61.3 & 45.3 \\
\hline
$\triangle$ Azimi et al.~\cite{Azimi_DOTA} & T+V & 68.2 & 81.4 & 74.3 & 47.7 & 70.3 & 64.9 & {\bf 67.8} & 70.0 & 90.8 & 79.1 & {\bf 78.2} & 53.6 & {\bf 62.9} & {\bf 67.0} & 64.2 & 50.2 \\
\hline
$\triangle$ ours & T & 67.4 & {\bf 88.3} & 71.7 & {\bf 51.4} & 66.5 & {\bf 72.4} & 64.5 & {\bf 76.7} & 90.8 & 77.3 & 74.2 & 45.9 & 60.2 & 65.7 & 56.7 & 48.3 \\
\hline
$\triangle$ ours & T+V & {\bf 69.9} & 87.8 & {\bf 82.4} & 49.4 & {\bf 73.5} & 71.1 & 63.5 & {\bf 76.7} & {\bf 90.9} & {\bf 79.2} & 73.3 & 48.4 & 60.9 & 62.0 & {\bf 67.0} & {\bf 62.2} \\
\hline
\end{tabular}}
\end{center}
\end{table*}

\begin{table*}[t]
\begin{center}
\small
\caption{Experiment Results of CAD-Net as well as Comparisons with State-of-the-art Techniques over the NWPU-VHR10 dataset.}
\label{tab:nwpuvhr10_results}
\begin{tabular}[t]{cccccccccccc}
\hline
{\bf method} & {\bf mAP} & {\bf PL} & {\bf ship} & {\bf ST} & {\bf BD} & {\bf TC} & {\bf BC} & {\bf GTF} & {\bf harbor} & {\bf bridge} & {\bf vehicle} \\
\hline
\hline
COPD~\cite{cheng2014multi} & 54.9 & 62.3 & 69.4 & 64.5 & 82.1 & 34.1 & 35.3 & 84.2 & 56.3 & 16.4 & 44.3  \\
\hline
Transferred CNN~\cite{alexnet} & 59.6 & 66.0 & 57.1 & 85.0 & 80.9 & 35.1 & 45.5 & 79.4 & 62.6 & 43.2 & 41.3  \\
\hline
RICNN~\cite{RSDD3} & 73.1 & 88.7 & 78.3 & 86.3 & 89.1 & 42.3 & 56.9 & 87.7 & 67.5 & 62.3 & 72.0  \\
\hline
Faster RCNN~\cite{FasterRCNN} & 84.5 & 90.9 & 86.3 & 90.5 & {\bf 98.2} & 89.7 & 69.6 & {\bf 100} & 80.1 & 61.5 & 78.1  \\
\hline
Li et al.~\cite{RotationInsensitiveAC} & 87.1 & {\bf 99.7} & {\bf 90.8} & 90.6 & 92.9 & {\bf 90.3} & 80.1 & 90.8 & 80.3 & 68.5 & 87.1  \\
\hline
ours (separation 1) & 92.1 & 90.9 & 80.8 & 96.4 & 90.9 & 90.2 & 90.9 & 99.6 & 100 & 90.9 & 89.9 \\
ours (separation 2) & 91.3 & 100 & 63.4 & 99.7 & 99.1 & 81.8 & 90.9 & 99.7 & 100 & 88.7 & 89.8 \\
ours (separation 3) & 91.0 & 100 & 89.6 & 90.6 & 90.8 & 90.8 & 79.6 & 99.5 & 100 & 79.0 & 90.0 \\
ours (Avg.)         & {\bf 91.5} & 97.0 & 77.9 & {\bf 95.6} & 93.6 & 87.6 & {\bf 87.1} & 99.6 & {\bf 100} & {\bf 86.2} & {\bf 89.9} \\
\hline
\end{tabular}
\end{center}
\end{table*}

\section{Experiments} \label{section:4}
This section presents experimentation including datasets and evaluation metrics, implementations details, experimental results over two public remote sensing object detection datasets, and ablation study of the proposed CAD-Net.

\subsection{Datasets and Evaluation Metrics}
The proposed technique is evaluated over two publicly available datasets as listed:

\smallskip
{\bf DOTA~\cite{DOTA}:} DOTA is a recently published large-scale open-access dataset for benchmarking object detection in remote sensing imagery. It is probably the largest and most diverse dataset on this task. It contains 2,806 aerial images that were captured using different sensors and platforms where over 188,000 object instances were annotated using quadrilaterals. The images from DOTA are diverse in sizes, ground sample distances (GSDs), sensor types, etc., and the captured objects also exhibit rich variation in term of scales, shapes and orientations. 15 categories of objects are annotated, which include plane (PL), baseball diamond (BD), bridge, ground track field (GTF), small vehicle (SV), large vehicle (LV), ship, tennis court (TC), basketball court (BC), storage tank (ST), soccer ball field (SBF), roundabout (RA), harbor, swimming pool (SP) and helicopter (HC). This dataset is divided into three subsets for training (1/2), validation (1/6), and test (1/3), respectively, where the ground truth of the test set is not publicly accessible.

\smallskip
{\bf NWPU-VHR10~\cite{cheng2014multi, cheng2016survey}:} NWPU-VHR10 is a publicly accessible dataset for object detection in remote sensing images. It has 800 very-high-resolution remote sensing images in total, among which 650 are positive and 150 are negative without containing any interested objects. The dataset has annotations of 10 types of objects including plane, ship, storage tank, baseball diamond, tennis court, basketball court, ground track field, harbor, bridge and vehicle. All interested objects are annotated using horizontal bounding boxes (HBB) that are publicly accessible.

\smallskip
{\bf Evaluation Metrics:}  We adopt the mean Average Precision (mAP) as the evaluation metric in all our experiments since mAP has been widely used for evaluation of multi-class object detection task in the literature. The definition of mAP is the same as the metric for PASCAL VOC 2012 object detection challenge.

\subsection{Implementation Details}
{\bf Ground Truth Generation:} DOTA provides annotations for objects of interest in quadrilateral format, while NWPU-VHR10 provides annotations in traditional axis-aligned bounding boxes format. To adapt to different settings, the proposed CAD-Net uses both horizontal bounding boxes (HBB) and oriented bounding boxes (OBB) as ground truth:
\begin{equation}
HBB: \left\{x_{min}, y_{min}, x_{max}, y_{max} \right\}
\end{equation}
\begin{equation} {\label{OBB_representation}}
OBB: \left\{x_{center}, y_{center}, w, h, \theta \right\}
\end{equation}
\noindent where $\theta$ lies within $[0, 90^\circ)$ to ensure that each object has a single ground truth. In training, the OBB ground truth as defined in Eq.(\ref{OBB_representation}) is generated by a set of rotated rectangles that best overlap with the provided quadrilateral annotations.

For DOTA dataset, our proposed CAD-Net generates both HBB results and OBB results, as is shown is Fig~\ref{fig:NetworkOverview}. For NWPU-VHR10 dataset, CAD-Net only generates HBB results, as OBB ground truth is not provided by this dataset.

\smallskip
{\bf Data Pre-processing:} Optical remote sensing images often have a huge image size, e.g. the size of DOTA images can be up to 6,000 $\times$ 6,000 pixels. To fit the hardware memory in training stage, we crop images into patches of size 1,600 $\times$ 1,600 pixels with an overlap of 800 pixels among neighboring patches. In the inference stage, image patches of 4,096 $\times$ 4,096 pixels are cropped from test images with an overlap of 1,024 pixels among neighboring patches. Zero padding is applied if an image is smaller than the cropped image patches. Other standard pre-processing processes are also performed such as global contrast normalization.

\smallskip
{\bf Network Setup:} We adopt the ResNet-101~\cite{Resnet} as network backbone for feature extraction. As a common practice, this ResNet-101 is pre-trained on the ImageNet~\cite{imagenet} and then fine-tuned over the training images of the two studied remote sensing image datasets during our training procedure. As objects in remote sensing images are often arbitrarily oriented, our proposed CAD-Net is designed to produce both HBB and OBB simultaneously as illustrated in Fig.~\ref{fig:NetworkOverview}, provided that the OBB ground truth is available. 

We adopt the stochastic gradient descent (SGD) with momentum for network optimization. Our model is trained on a single Nvidia Tesla P100 SXM2 GPU with 16GB memory, along with the deep learning framework PyTorch~\cite{PyTorch}. Batch size is set to 1. Total training iterations for DOTA and NWPU-VHR10 are 130,000 and 30,000, which take around 36 and 6 hours, respectively.

\begin{figure*}[t]
\begin{center}
   \includegraphics[height=122mm, width=1\linewidth]{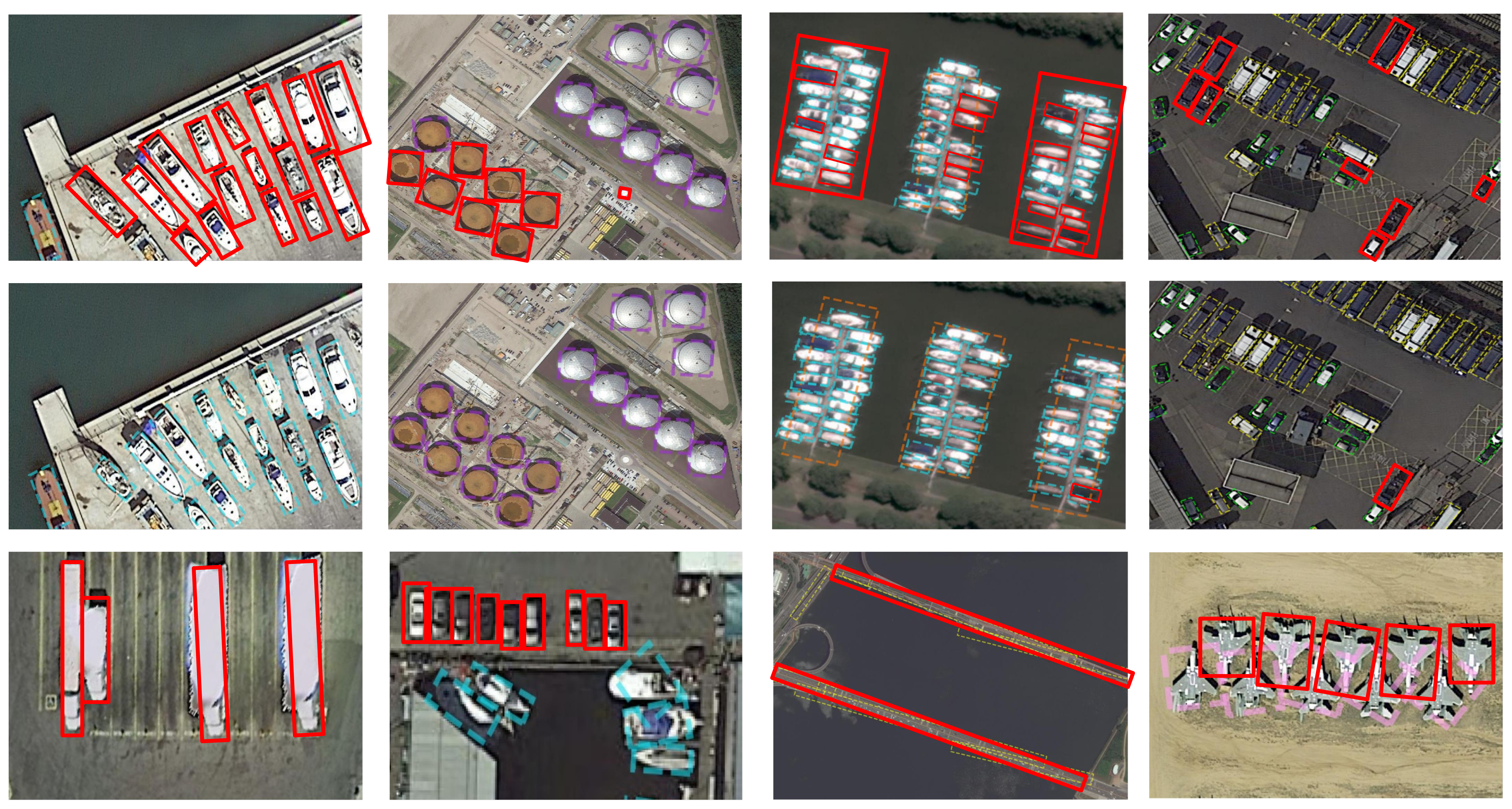}
\end{center}
   \caption{Illustration of object detection in remote sensing images by our proposed CAD-Net: The sample images in the first row show detection by the baseline Faster R-CNN with FPN. The second row show detection by our proposed CAD-Net where most objects are better detected correctly. The sample images in the third row show typical failure cases by our proposed CAD-Net, where dashed boxes of different colors show True Positive detection of objects of different categories and red-color boxes denote False Positives or False Negatives. Qualitative experiment shows that our proposed CAD-Net is tolerant to different types of degradation and information loss (all sample images are from DOTA~\cite{DOTA} dataset).}
\label{fig:DetResult}
\end{figure*}

\subsection{Experimental Results} 
Table \ref{tab:dota_results} shows experimental results on the {\bf test set} of DOTA dataset and comparisons with state-of-the-art methods. Note that all methods listed in this table adopt ResNet-101 as backbone network, except that YOLO v2 and SSD which adopt GoogLeNet and Inception network respectively.  As Table \ref{tab:dota_results} shows, our proposed CAD-Net outperforms the Faster RCNN baseline (FR-O in Table \ref{tab:dota_results}) ~\cite{DOTA} by a large margin (+ 15.8\% mAP), demonstrating its effectiveness for object detection from remote sensing images. In addition, it outperforms state-of-the-art performance by up to 2\% under two training setups (`T' means only training images are used in training and `T+V' means both training images and validation images are used in training). Besides, we would indicate that Azimi's method adopts Inception module~\cite{inception}, deformable convolution~\cite{defconv}, online hard example mining (OHEM)~\cite{OHEM}, multi-scale training and inference, etc., whereas we target a clean and efficient model with outstanding performance. Our model should be able to achieve higher detection accuracy by including those well-proven performance-boosting components.

We also evaluate the proposed CAD-Net on the NWPU-VHR10 dataset and benchmark it with state-of-the-art methods. Since NWPU-VHR10 does not specify the partition of training and testing sets, we randomly choose 75\% of the positive images as training set and the rest positive images as test set by following the widely adopted partition scheme~\cite{RotationInsensitiveAC}, except that we do not include any negative images for training. Table~\ref{tab:nwpuvhr10_results} shows experimental results and comparisons with state-of-the-art method. We provide experimental results of CAD-Net on 3 random separations of NWPU-VHR10 dataset to provide more compelling results. As Table~\ref{tab:nwpuvhr10_results} shows, the proposed CAD-Net also obtain superior object detection performance as compared with state-of-the-art methods.

Fig.~\ref{fig:DetResult} shows a few sample images from the DOTA dataset and the corresponding detection by using the baseline model--Faster RCNN with FPN (in the first row) and the proposed CAD-Net (in the second row). As Fig.~\ref{fig:DetResult} shows, the state-of-the-art generic detection technique Faster RCNN with FPN tends to produce incorrect detection under different scenarios such as ships in the first sample image (mis-detection as large vehicles), storage tanks of different styles in the second sample image (mis-detection as roundabouts), harbors occluded by ships and ships with little texture detail in the third sample image (false negatives) and vehicles with very low contrast with the background in the fourth sample image (false negatives). As a comparison, the proposed CAD-Net is capable of correctly detecting those objects under various adverse scenarios as illustrated in the second row of Fig.~\ref{fig:DetResult}. The superior detection performance is largely attributed to the inclusion of the global contexts, local contexts, spatial-and-scale-aware attention, strong and balanced semantics information and accurate rotation angle regression (as described in Section III) within the proposed CAD-Net.

On the other hand, the proposed CAD-Net are still prone to detection failures under several typical situations as illustrated in the third row of Fig.~\ref{fig:DetResult}. First, the proposed CAD-Net is sensitive to strong light interference as illustrated in the first sample image, largely due to the lack of relevant training images within the training set. Second, CAD-Net often produces missed detection for small vehicles even when the small vehicles have good visual quality as illustrated in the second sample image. We strongly believe that this is largely due to the inaccurate annotation of the training images. In particular, many small vehicles are not annotated, probably because of a huge amount of small vehicles in images and limited manpower. Third, CAD-Net may fail to detect long and thin objects such as bridges as shown in the third sample image. This is a common constraint of proposal-based detection techniques such as Faster RCNN which can only employ a limited number of anchors for objects with limited aspect ratios. Fourth, CAD-Net still tends to miss detecting objects that are heavily overlapped with each other as shown in the fourth sample image. We believe this issue can be better addressed by proper Non-Maximum-Suppression (NMS), which we will investigate in our future work. 

\subsection{Ablation Study} \label{section:ablationstudy}
We perform an ablation study to identify the contributions of the proposed GCNet, PLCNet and spatial-and-scale-aware attention over the DOTA dataset. Several network models are trained for the ablation study which mainly include: {\bf Baseline:} The network in beige in Fig.~\ref{fig:NetworkOverview} which is actually a Faster RCNN with FPN and ResNet-101 network backbone (we follow all configurations as~\cite{FPN}); {\bf Baseline + GCNet:} The Baseline with GCNet included only; {\bf Baseline + PLCNet:} The Baseline with PLCNet included only, where PLCNet is built upon FPN-generated feature pyramid since attention modulated feature pyramid does not exist in this setting; {\bf Baseline + Spatial-Scale-Aware Attention:} The Baseline with spatial-and-scale-aware attention included only; and {\bf CAD-Net:} The full implementation of the proposed context-aware detection network as shown in Fig.~\ref{fig:NetworkOverview}. Experiments for paired components are also included to verify the complementarity of the proposed components.

\begin{table}[t]
\renewcommand{\arraystretch}{1.3}
\caption{Ablation Study of the proposed CAD-Net over the DOTA Validation Set.}
\label{tab:ablationstudy}
\centering
\begin{tabular}[t]{cc}
\hline
{\bf setting} & {\bf mAP (\%)} \\\hline\hline
Baseline (Faster RCNN with FPN) & 59.8  \\\hline
Baseline + GCNet & 62.4 \\\hline
Baseline + PLCNet & 61.2 \\\hline 
Baseline + Spatial-Scale-Aware Attention & 62.4 \\\hline
Baseline + GCNet + PLCNet & 62.9 \\\hline
Baseline + GCNet + Attention & 63.8 \\\hline
Baseline + PLCNet + Attention & 63.9 \\\hline
CAD-Net & 64.8 \\\hline
\end{tabular}
\end{table}

Table \ref{tab:ablationstudy} shows experimental results over the {\bf validation set} of the DOTA dataset. As Table \ref{tab:ablationstudy} shows, the model \textbf{Baseline} can only achieve a mAP of 59.8\%. By including the GCNet, the model \textbf{Baseline + GCNet} achieves a $\sim$2.5\% mAP improvement, demonstrating the effectiveness of including global scene-level contextual information. The model \textbf{Baseline + PLCNet} also achieves a $\sim$1.5\% mAP improvement while the local contextual information is included, demonstrating the effectiveness of including neighboring objects/features in object detection in aerial imagery. The model \textbf{Baseline + Spatial-Scale-Aware Attention} achieves a similar $\sim$2.5\% mAP improvement when the proposed attention module is included, demonstrating the effectiveness of including our proposed attention module to generate spatial-and-scale-aware feature maps. In addition, three experiments with different paired components are included to demonstrate that the proposed PLCNet, GCNet and attention module are actually complementary to each other. Finally, the proposed CAD-Net that combines the GCNet, PLCNet and attention module achieves a 5\% mAP improvement compared to the \textbf{Baseline} model, pushing the mAP to 64.8\%. The results of this ablation study well align with our motivations.

Note that the ground truth annotations of the test set of DOTA are not publicly accessible, and the number of submissions for evaluation on the test set is also limited by the dataset creators. Therefore, all experiments of this ablation study are therefore evaluated over the DOTA validation set which provides publicly accessible object annotations.
 
\section{Conclusions}
This paper presents a novel CAD-Net, an accurate and robust detection network for objects in optical remotes sensing images. Global Context Network (GCNet) and Pyramid Local Context Network (PLCNet) are proposed, which extract scene-level and object-level contextual information that is highly correlated to objects of interest and often provide extra guidance for object detection in remote sensing images. In addition, a spatial-and-scale-aware attention module is designed which guides the network to focus on scale-adaptive features for feature maps from each level and also to emphasize the degraded texture details. Extensive experiments over two public available datasets verify the uniqueness of object detection in remote sensing images, and also show that the proposed CAD-Net achieves superior object detection performance as compared with state-of-the-art techniques. On the other hand, the CAD-Net still tends to fail under several typical scenarios for ultra-long or heavily overlapped objects. We will investigate new approaches that is capable of better leveraging contextual information for more robust object detection in remote sensing images.


\ifCLASSOPTIONcaptionsoff
  \newpage
\fi

\bibliographystyle{IEEEtran}
\bibliography{IEEEabrv,ref}

%

\begin{IEEEbiography}[{\includegraphics[width=1in,height=1.25in,clip,keepaspectratio]{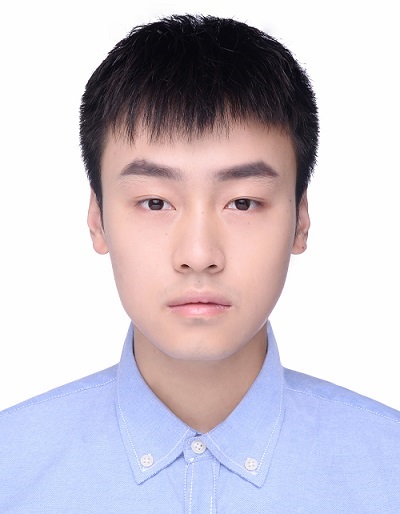}}]{Gongjie Zhang}
Mr. Gongjie Zhang is currently working toward the PhD degree in the Nanyang Technological University, Singapore, under the supervision of Dr. Shijian Lu. He received his B.Eng. degree in electronic and information engineering in 2018 from Northeastern University, Shenyang, China. His research interests mainly include computer vision, object detection, satellite imagery analytics and meta-learning.
\end{IEEEbiography}

\begin{IEEEbiography}[{\includegraphics[width=1in,height=1.25in,clip,keepaspectratio]{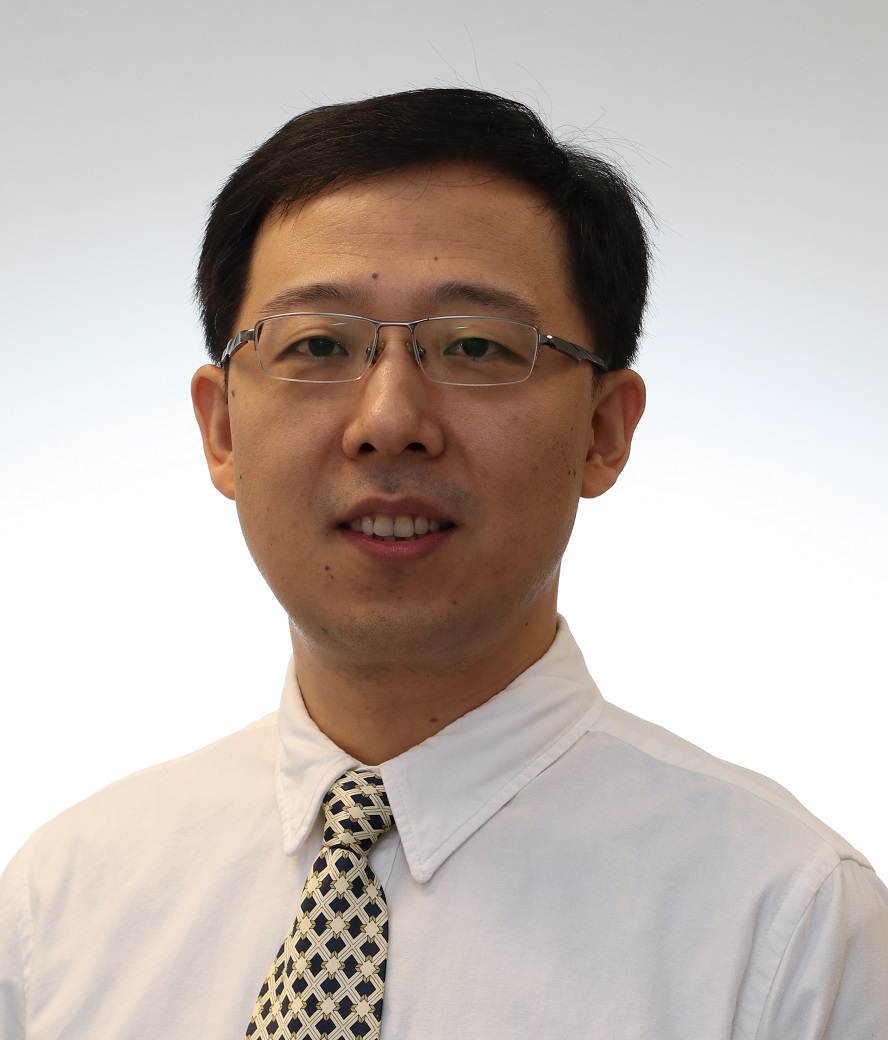}}]{Shijian Lu}
Dr. Shijian Lu received his PhD in electrical and computer engineering from the National University of Singapore. He is an Assistant Professor with School of Computer Science and Engineering, the Nanyang Technological University, Singapore. His major research interests include image and video analytics, visual intelligence, and machine learning. He has published more than 100 international refereed journal and conference papers and co-authored over 10 patents in these research areas. He is currently an Associate Editor for the journal Pattern Recognition (PR). He has also served in the program committee of a number of conferences, e.g. the Area Chair of the International Conference on Document Analysis and Recognition (ICDAR) 2017 and 2019, the Senior Program Committee of the International Joint Conferences on Artificial Intelligence (IJCAI) 2018 and 2019, etc.
\end{IEEEbiography}

\begin{IEEEbiography}[{\includegraphics[width=1in,height=1.25in,clip,keepaspectratio]{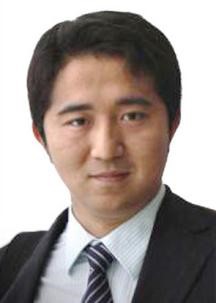}}]{Wei Zhang}
Wei Zhang (S'06-M'11) received the Ph.D. degree in Electronic Engineering from The Chinese University of Hong Kong in 2010. He is currently a Professor of the School of Control Science and Engineering at Shandong University, China. His research interests include computer vision, image processing, pattern recognition, and robotics. He has published over 70 papers in international journals and refereed conferences. He served as a program committee member and reviewer for various international conferences and journals in image processing, computer vision and robotics.
\end{IEEEbiography}




\end{document}